%% file: main.tex
\documentclass[letterpaper, 10 pt, conference]{ieeeconf}
\IEEEoverridecommandlockouts
\let\labelindent\relax
\usepackage{enumitem}
\usepackage{balance}
\usepackage{array}
\usepackage{textcomp}
\usepackage{mathtools, nccmath}
\usepackage{graphicx}
\usepackage{subfigure}
\usepackage{amsfonts}
\usepackage{amsmath}
\usepackage{amssymb}
\usepackage{algorithm}
\usepackage{algorithmic}
\usepackage{tikz}
\usepackage{arydshln}
\usepackage{multirow}
\usepackage{bm}
\usepackage{epstopdf}
\usepackage{siunitx}
\usepackage{xcolor}
\usepackage{cases}
\usepackage{cite}
\makeatletter
\let\NAT@parse\undefined
\makeatother
\usepackage[colorlinks=true,
            linkcolor=blue,
            citecolor=green
           ]{hyperref}

\usepackage{glossaries}
    \glsdisablehyper 

\input{macros.tex}

\input{gloss.tex}

\title{\LARGE \bf Collaborative Trolley Transportation System with \\Autonomous Nonholonomic Robots}
\author{
        Bingyi~Xia$^\dag$, 
        Hao~Luan$^\dag$, 
        Ziqi~Zhao$^\dag$, 
        Xuheng~Gao, 
        Peijia~Xie,
        Anxing~Xiao, \\
        Jiankun~Wang, 
        \IEEEmembership{Senior~Member,~IEEE}, 
        and Max~Q.-H.~Meng, 
        \IEEEmembership{Fellow,~IEEE}
        \thanks{$^\dag$Equal contribution. }
        \thanks{This work is partially supported by Shenzhen Key Laboratory of Robotics Perception and Intelligence (ZDSYS20200810171800001), National Natural Science Foundation of China grant \#62103181, and Shenzhen Outstanding Scientific and Technological Innovation Talents Training Project under Grant RCBS20221008093305007. \emph{(Corresponding authors: Jiankun Wang, Max~Q.-H.~Meng.)}}
        \thanks{Bingyi Xia, Ziqi Zhao, Xuheng Gao, Peijia Xie, Jiankun Wang and Max Q.-H. Meng are with Shenzhen Key Laboratory of Robotics Perception and Intelligence, Department of Electronic and Electrical Engineering, Southern University of Science and Technology, Shenzhen, China. {\tt \small\{xiaby2020, zhaozq2020, gaoxh2021, xiepj2022\}@mail.sustech.edu.cn}, {\tt \small wangjk@sustech.edu.cn}, {\tt \small  max.meng@ieee.org}}
        \thanks{Jiankun Wang is also with the Jiaxing Research Institute, Southern University of Science and Technology, Jiaxing, China.}
        \thanks{Max~Q.-H.~Meng is also a Professor Emeritus in the Department of Electronic Engineering at the Chinese University of Hong Kong in Hong Kong and was a Professor in the Department of Electrical and Computer Engineering at the University of Alberta in Canada.}
        \thanks{Hao Luan and Anxing Xiao are with the School of Computing, National University of Singapore, Singapore.
        {\tt\small \{haoluan, anxingx \}@comp.nus.edu.sg}
 }
    }

\begin{document}
\maketitle

\begin{abstract}
Cooperative object transportation using multiple robots has been intensively studied in the control and robotics literature, but most approaches are either only applicable to omnidirectional robots or lack a complete navigation and decision-making framework that operates in real time.
This paper presents an autonomous nonholonomic multi-robot system and an end-to-end hierarchical autonomy framework for collaborative luggage trolley transportation. 
This framework finds kinematic-feasible paths, computes online motion plans, and provides feedback that enables the multi-robot system to handle long lines of luggage trolleys and navigate obstacles and pedestrians while dealing with multiple inherently complex and coupled constraints.
We demonstrate the designed collaborative trolley transportation system through practical transportation tasks, and the experiment results reveal their effectiveness and reliability in complex and dynamic environments. 
(Video\footnote{Video demonstration: \url{https://youtu.be/efnPERm0Rco}.})
\end{abstract}

\IEEEpeerreviewmaketitle

\input{sections/1_introduction.tex}
\input{sections/2_related_works.tex}
\input{sections/3_system.tex}
\input{sections/4_method.tex}

\input{sections/5_exp.tex}

\input{sections/6_conclusion.tex}

{
    \bibliographystyle{IEEEtran}
    \bibliography{bib/bibliography, bib/hl}
}

\end{document}

%% file: macros.tex
\newcommand{\figref}[1]{Fig.~\ref{#1}}
\newcommand{\secref}[1]{Section~\ref{#1}}
 
\newcommand{\apriori}{\emph{a priori }}

\newcommand{\etc}{\textit{etc. }}

\newcommand{\ie}{\textit{i.e.}}
\DeclareMathOperator{\arctantwo}{arctan2}
\newcommand{\mvc}[1]{\mathbf{#1}}

\newcommand{\matSym}[1]{\boldsymbol{#1}}

\newcommand{\T}{^\top} 

\newcommand{\dtSample}{\Delta t}

\newtheorem{rmk}{Remark}


%% file: gloss.tex
\setacronymstyle{long-short}

\newacronym{roi}{ROI}{region of interest}
\newacronym{mpc}{MPC}{model predictive control}
\newacronym{nmpc}{NMPC}{nonlinear model predictive control}
\newacronym{prm}{PRM}{Probablistic RoadMap}
\newacronym{rrt}{RRT}{Rapidly Random-exploring Tree}
\newacronym{rrt*}{RRT$^*$}{}
\newacronym{cbf}{CBF}{control barrier function}
\newacronym{clf}{CLF}{control Lyapunov function}
\newacronym{dtcbf}{DT-CBF}{discrete-time control barrier function}
\newacronym{dtclf}{DT-CLF}{discrete-time control Lyapunov function}
\newacronym{clfcbfqp}{CLF-CBF-QP}{CLF-CBF-QP}
\newacronym{dclf-dcbf}{DCLF-DCBF}{}
\newacronym{ekf}{EKF}{extended Kalman filter}

\newacronym[
    longplural={Nash equilibria},
    shortplural={NE}
]{ne}{NE}{Nash equilibrium} 
\newacronym[
    longplural={finite-state automata},
    shortplural={FSA}
]{fsa}{FSA}{finite-state automaton} 


%% file: sections/1_introduction.tex
\section{Introduction}%
\label{sec:introduction}

Robots are versatile tools for object manipulation and transportation\cite{tang2022relationship}, and have a broad range of applications, including industry assembly lines \cite{johannsmeier2016hierarchical}, vehicle extraction \cite{amanatiadis2015avert}, and luggage collection at airports \cite{wang2021real,xiao2022robotic}, \etc 
In many cases, the movement of large objects requires the coordination of multiple robots for enhanced strength or mobility.
Transporting a chain of collected luggage trolleys in complex and congested environments such as international airports exemplifies such an application. 
This task is challenging due to several factors: 
\romannumeral1) the robots have to carry a long stack of omnidirectional trolleys, 
\romannumeral2) the operating environment may be densely populated and with narrow corridors and tight corners, and 
\romannumeral3) the robots have to perform highly coordinated movements to maintain the integrity of the trolley stack.

In this paper, we present a practical multi-robot system along with a hierarchical navigation framework for the task of transporting a series of luggage trolleys with autonomous robots.
Two nonholonomic robots that were previously used in our trolley collection work \cite{xiao2022robotic} are further adapted and form a robot team for this transportation task. 
To tackle the aforementioned challenges, we propose an end-to-end pipeline consisting of several modules that address perception, behavioral planning, global pathfinding, and collaborative motion planning. 
The proposed framework enables our robot team to work collaboratively, transport a long stack of luggage trolleys, and safely navigate complex and dynamic environments in real time.

In this work, our contributions are threefold:
\begin{itemize}
    \item
    We develop a practical nonholonomic multi-robot system for the collaborative transportation of a series of luggage trolleys. 
    \item
    We propose a hierarchical real-time planning framework for the safe navigation of a nonholonomic multi-robot system, addressing intricate constraints with tight inter-robot motion couplings. 
    \item
    We demonstrate that the designed system can achieve collaborative luggage trolley transportation in complex and dynamic real-world environments. 
\end{itemize}


\begin{figure}[t]
    \centering
    \includegraphics[width=.98\linewidth]{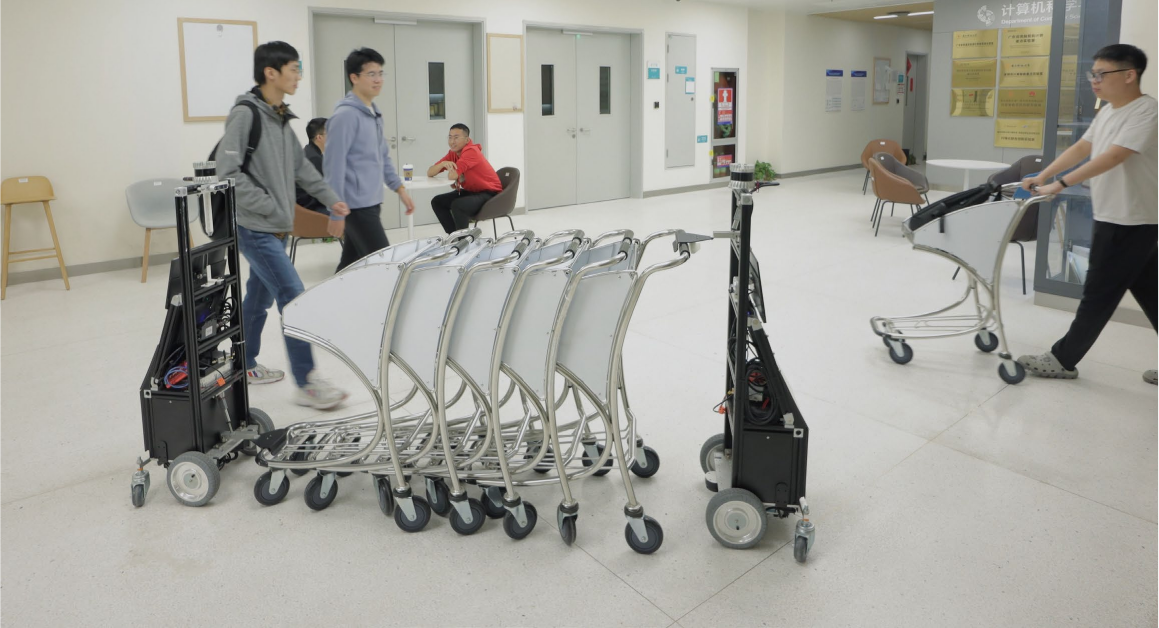}
    \caption{
        A snapshot showing a dual-robot system transporting a series of luggage trolleys in tandem in a dynamic environment.
    }
    \label{fig:superimposed_snapshots}
    \vspace{-0.6cm}
\end{figure}

\begin{figure*}[tb]
    \centering
    \includegraphics[width=.99\linewidth]{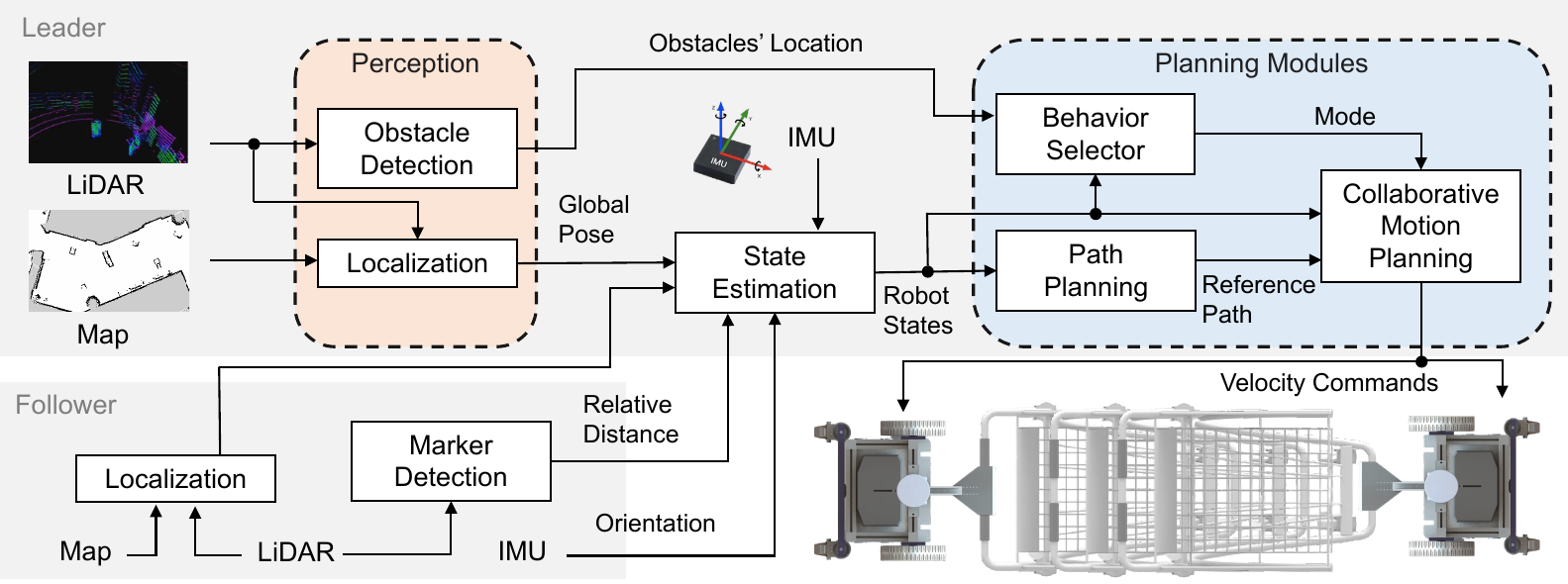}
    \caption{
        A diagram of our proposed navigation framework. 
        It starts by measuring different sensory information, which is then processed by the perception module. 
        Thereafter, the planning modules perform real-time computation and provide each robot with its velocity command. 
    }
    \label{fig:framework}
    \vspace{-5mm}
\end{figure*}  

%% file: sections/2_related_works.tex
\section{Related Work}%
\label{sec:related_works}


Cooperative manipulation and transportation have been traditionally approached as a control task, where the desired trajectories of manipulated objects are often predefined\cite{2018Cooperative}. 
The robots need to compute control inputs in a central or distributed fashion to enable the manipulated objects to track the desired trajectories or paths. 
A decentralized control framework is brought forth in \cite{sugar2002control}, enabling robotic collaborative manipulation and transportation of large objects via separating locomotion and grasping. 
A number of works opt for the caging method \cite{wang2002object, pereira2004decentralized}, in which the manipulated object stays in the interior of the formations formed by the robots. 
Without any explicit communication and with merely local measurements, the approach along with the physical system proffered in \cite{ wang2018ouijabots} realizes multi-robot cooperative object transportation via force coordination under a leader-follower paradigm. 
In \cite{culbertson2021decentralized}, the authors provide a distributed adaptive control algorithm for several omnidirectional robots manipulating a rigid body to track a desired trajectory in $SE(3)$.  
However, these works all assume \apriori known reference paths or trajectories and do not provide navigation solutions in dynamic environments. 
Moreover, control algorithms in \cite{wang2002object, pereira2004decentralized, wang2018ouijabots,culbertson2021decentralized} are all designed for omnidirectional robots without nonholonomic constraints, making them unsuitable for many practical scenarios, such as the luggage trolley transportation work mentioned in this paper. 

The advancement of onboard computing power has unlocked real-time deployment of optimization-based robot planning and control methods like \gls{mpc}. 
In \cite{alonso-mora2015local}, the authors formulate the local control of manipulating deformable objects as a convex optimization problem and propose to solve it in a receding-horizon fashion with a centralized/distributed planning approach. 
Many studies have also investigated aerial cooperative payload carrying\cite{tagliabue2019robust, jackson2020scalable} and connected vehicles\cite{liu2021securing, zhu2021safety}. 
As expected, centralized optimization scales poorly with the number of robots. 
Thereupon, decentralized trajectory optimization methods keep drawing the field's attention. 
The authors of \cite{ebel2021design} provide us with a decentralized \gls{mpc} planning and control scheme for moving polygon objects with a set of omnidirectional robots on a plane.
Though a distributed algorithm for optimization with separable variables is leveraged in multi-robot manipulation \cite{shorinwa2020scalable}, it is not well-suited for multi-agent systems with highly nonlinear dynamics and strong couplings. 
Recent study have explored holonomically constrained collaborative locomotion \cite{kim2022layered} and their use in collaborative manipulation of cable-towed loads \cite{yang2022collaborative}.  
Nevertheless, none of the aforementioned approaches fits our task, which features tight coupling and complex constraints between robots, nonholonomic dynamics, and a demand for high motion precision. 

Previous collaborative object transportation work in similar settings to ours has mainly focused on control.
Yufka and Ozkan~\cite{yufka2015formationbased} take a formation-control perspective and propose a virtual leader-follower control scheme enabling multiple differential-drive robots to transport a pallet-like object. 
The work in \cite{tsiamis2015decentralized} investigates a decentralized control algorithm for two nonholonomic robots without direct communication and provides thorough theoretic analyses. 
The authors of \cite{yasuda2021cooperative} employ risk-sensitive stochastic control for the same task, but only use this approach to compute one control input for a bi-robot system. 
Works in \cite{yufka2015formationbased,tsiamis2015decentralized,yasuda2021cooperative} all lack a systematic navigation solution that is essential for robots to perform the transportation task in complex, dynamic, and uncertain environments. 
The closest work to ours is, perhaps, \cite{machado2016multiconstrained}. 
The proposed approach allows for the integration of tracking control and obstacle avoidance. 
Nonetheless, the proposed method involves multiple handcrafted attractor functions and quite a few parameters that entail substantial fine tuning. 
In contrast, our hierarchical approach in this paper features algorithmic flexibility at different decision-making levels, and the parameter tuning is more intuitive and practical. 

%% file: sections/3_system.tex
\section{System Description}
\label{sec:system_description}

\subsection{Framework Overview}%
\label{sec:the_overall_framework}

We introduce an end-to-end autonomy framework for robotic collaborative trolley transportation in complex and dynamic environments. 
Our framework is illustrated in \figref{fig:framework}. 
For perception, multiple sensors are equipped to acquire both global and relative localization measurements, taking uncertainty into account. 
The proposed hierarchical decision-making module comprises three parts: 
\romannumeral1) the global path planner, 
\romannumeral2) the behavior selector, and 
\romannumeral3) the collaborative motion planner. 

The three planning parts are built for three core objectives that ensure the safety and reliability of the trolley transportation task: kinematic-feasible trajectory generation, obstacle avoidance, and trolley array integrity maintenance. 
Starting with a 2D occupancy grid map, the global path planner provides a sequence of navigation waypoints by treating the entire robot-trolley assembly as a single virtual vehicle and considering kinematic constraints.
Then, the collaborative motion planner computes online the two robots' trajectories and velocity commands that track the reference path fed by the path planner and accounts for a variety of constraints, including nonholonomic dynamics, coupling formation constraints, and control bounds. 
The behavior selector evaluates the risk of collision with detected dynamic obstacles like pedestrians nearby, and adaptively adjusts the parameters of the motion planner or directly modifies the velocity commands if necessary. 
The advantage of such decomposition primarily resides in guaranteeing real-time performance for solving the complex planning and control problem with tight inter-agent couplings while still delivering precise and highly coordinated motions.

\subsection{Mechanical Design}
\label{sub:design}

 \begin{figure}[tb]
    \centering
    \includegraphics[width=.95\linewidth]{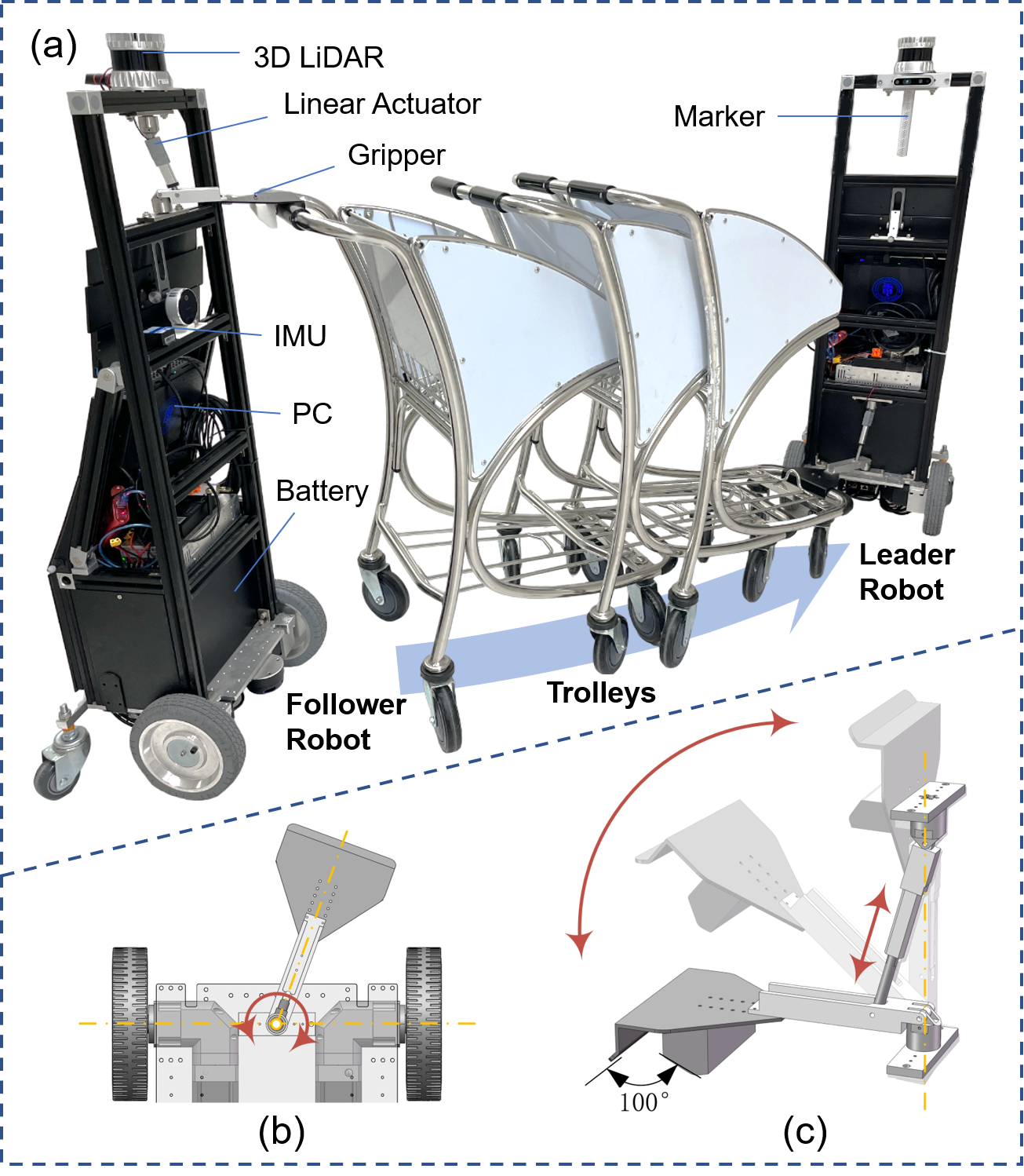}
    \caption{
    A prototype of the trolley collaborative transportation system is shown in \figref{fig:design}(a), with the robot components labeled. \figref{fig:design}(b) and (c) illustrate the structure and kinematics of the manipulator.}
    \label{fig:design}
    \vspace{-6mm}
\end{figure}

We develop a multi-robot system consisting of a leader and a follower for the collaborative trolley transportation task.

    \subsubsection{Components}
    \label{ssub:Components}
    Adapted from the previous generation of luggage trolley collection robots \cite{xiao2022robotic}, the leader and follower share a similar compact structure and identical components, including a differential-drive chassis with suspension, an aluminum alloy frame, large-capacity batteries, wiring, multiple sensors (LiDARs, cameras, IMU, \etc), an onboard computer, and a manipulator, as shown in \figref{fig:design}(a).

    \subsubsection{Manipulator Design}
    \label{ssub:grasp}
    The manipulator comprises a linear actuator and a gripper, which use leverage to grasp a trolley's handler or beam, as depicted in \figref{fig:design}(c). 
    The manipulator's placement is adjustable and dependent on the height of the handler or beam being gripped. 
    To ensure that the gripper grasps the beam at the proper horizontal angle, the bottom of the manipulator must be at the same height as the beam. 
    This placement provides a secure grip and acts as protection preventing axial forces on the linear actuator during transportation. 
    As is shown in \figref{fig:design}(a), the leader robot's manipulator is at a low position gripping the front beam of the trolley, while the follower robot's manipulator is positioned high to grip the handler. 

    To enable the two robots to transport trolleys collaboratively along curved paths, we employ a passive joint connecting the gripper and the robot, as shown in \figref{fig:design}(b). 
    By aligning the rotation axis of the gripper with that of the chassis, we eliminate the torque caused by the force exerted by the trolley queue on the gripper. 
    This design effectively reduces the impact of the robots' angular error and the dynamic adjustment required by the collaborative trolley transport system.

%% file: sections/4_method.tex
\section{Methodology}%
\label{sec:methodology}

In this section, we elaborate our hierarchical decision-making framework in three parts: the collaborative motion planner, the global path planner, and the behavior selector. 
In addition, we introduce how we fuse information from multiple sources for reliable state estimation.

\subsection{Collaborative Motion Planning}%
\label{sub:motion_planning}
The main challenge of the collaborative transportation task lies in the collaborative motion planning of the robot team. 
The robots need to compute their control inputs such that the robot-trolley assembly follows a desired path yielded by the global path planning module and meanwhile, to keep the trolleys sticking together and firmly attached to the robots. 
These requirements involve multiple constraints over the local collaborative motion planning problem. 
Therefore, we opt for a \gls{nmpc} based approach. 
The \gls{nmpc} framework provides great convenience for posing constraints, and has been successfully used in many robotic applications \cite{xiao2021robotic,jian2022putn}. 
In this section, we first describe the mathematical model of the robot-trolley system and then formulate an optimization problem for collaborative transportation motion planning.

\begin{figure}[t]
    \centering
    \includegraphics[width=.85\linewidth]{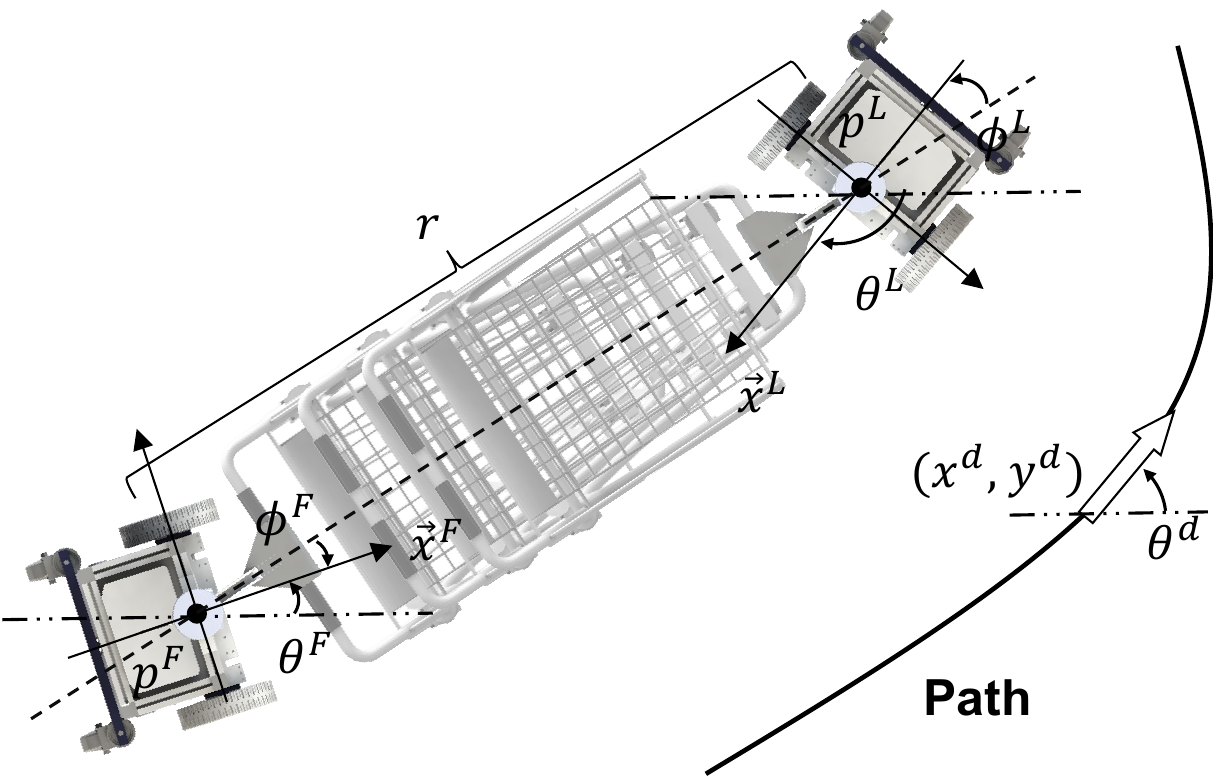}
    \caption{
        Illustration of the robot-trolley assembly tracking a reference path. The trolley stack is assumed to be one single rigid body.
    }
    \label{fig:lfmodel}
    \vspace{-6mm}
\end{figure}

The robot-trolley assembly consists of two differential-drive robots and an array of omnidirectional trolleys. 
Specifically, one robot $L$, a leader, is in charge of steering the trolleys at the head of the array, and a follower robot $F$ provides power and persists the formation configuration at the tail, as shown in \figref{fig:lfmodel}. 
Assuming the trolleys are properly attached and the number of trolleys is known, we use the two robots' planar positions and heading angles as the assembly system's state and take the robots' linear and angular velocities as the system's input: 
\begin{align}
    \mvc{x} &\triangleq 
        \begin{bmatrix}
        x^L & y^L & \theta^L & x^F  & y^F & \theta^F
        \end{bmatrix}\T \in \mathbb{R}^6  ,\\ 
    \mvc{u} &\triangleq 
        \begin{bmatrix}
        v^L & \omega^L & v^F & \omega^F 
        \end{bmatrix}\T \in \mathbb{R}^4 ,
\end{align}
in which $\{L, F\}$ represents the role (``Leader'' or ``Follower'') of each robot. 
Thereupon, the assembly's kinematics is formulated as two concatenated discrete-time unicycle model: 
\begin{equation}
\label{eq:dynamics}
    \mathbf{x}_{k+1} = f(\mathbf{x}_k, \mathbf{u}_k) = \mathbf{x}_k + \matSym{G}(\mathbf{x}_k) \mathbf{u}_k ,
\end{equation}
where 
\begin{equation*}
    \matSym{G}(\mathbf x_k) \triangleq \dtSample
        \begin{bmatrix}
            \cos\theta^L_k & 0 & 0 & 0 \\
            \sin\theta^L_k & 0 & 0 & 0 \\
            0 & 1 & 0 & 0 \\
            0 & 0 & \cos\theta^F_k & 0 \\
            0 & 0 & \sin\theta^F_k & 0 \\
            0 & 0 & 0 & 1 \\
        \end{bmatrix}
,\end{equation*}
and $\dtSample$ is the sampling time interval. 
Herein, we further denote a number of variables derived from the state. 
The relative distance between the two robots is of great interest: 
\begin{equation}
\label{eq:rel_dist}
    r \triangleq \| \mvc{p}^L - \mvc{p}^F \|,
\end{equation}
where $\mvc{p}^i \in \mathbb{R}^2$ for $i\in \{L,F\}$ is the planar position of a robot. 
Moreover, we denote the angle difference between the orientation of each robot and that of the vector that starts from the follower pointing at the leader as
\begin{equation}
    \phi^i \triangleq \theta^i - \arctantwo{\left( y^L-y^F, x^L-x^F \right)}, \quad i \in \{L, F\}
.\end{equation}
\figref{fig:lfmodel} illustrates the geometrical relationships among these variables.
\begin{rmk}
The desired relative distance $l$ between the two robots is a mapping $\kappa:\, N \to l$ with respect to the number of trolley(s) $N$.
This quantity is assumed to be known \apriori from previous trolley collection tasks, so it is a constant here. 
\end{rmk}
\begin{rmk}
As shown in \figref{fig:lfmodel}, the center of the trolley stack does not necessarily coincide with the center of the line connecting the two robots. 
Under the condition that the trolleys are tightly arranged and the manipulators passively keep the same heading angle with the trolleys, we find that this slipping between manipulators and trolleys is negligible for motion planning in practice.
\end{rmk}

Now we pose the collaborative motion planning problem as an \gls{nmpc} problem as follows:
\begin{subequations}
\label{eq:opti}
\begin{align}
    \min_{\{\mvc{x}, \mvc{u}, \varepsilon\}} \quad& J_{\text{T}}(\mvc{x}_T) + \sum^{T-1}_{k=0} J_{\text{S}}(\mvc{x}_k, \mvc{u}_k, \varepsilon_k) \label{eq:opti_cost} \\
    \text { s.t. }\quad & \mvc{x}_{k+1} = f(\mvc{x}_k, \mvc{u}_k) \label{eq:opti_dyn}\\
    &\mvc{x}_0= \mvc{x}_{\text{init}} \label{eq:opti_init}\\
    &|r_k - l| \le \varepsilon_k \label{eq:opti_dist}\\
    &\mvc{x}_k \in \mathcal{X} \label{eq:opti_x_feas}\\
    &\mvc{u}_k \in \mathcal{U} \label{eq:opti_u_feas},
\end{align}
\end{subequations}
where $J_{\text T}$ is the terminal cost, $J_{\text S}$ is the stage cost, and $\varepsilon_k$ is a slack variable.

The constraints are, from top to bottom, respectively: 
\eqref{eq:opti_dyn} the kinematics constraint from \eqref{eq:dynamics}; 
\eqref{eq:opti_init} initial conditions; 
\eqref{eq:opti_dist} the relative distance constraint, relaxed by a slack variable to avoid infeasibility; 
\eqref{eq:opti_x_feas} workspace constraints; 
\eqref{eq:opti_u_feas} admissible control constraints, in which the robots' velocities are bounded in the sense of $|v^i| \le v_{\textrm{max}}^i$, and $|\omega_i| \le \omega_{\textrm{max}}^i$, and accelerations are bounded via discretization as well $|{v}^i_{k+1} - v^i_k| \le a^i_{\text{max}}$, $ |\omega^i_{k+1}-\omega^i_{k}| \le \alpha^i_{\text{max}}$. 

The objective function \eqref{eq:opti_cost} contains several quadratic costs for path tracking and formation maintenance.  
Path tracking is achieved by employing a pure pursuit strategy encoded in the terminal cost: 
\begin{equation}
    J_{\text{T}}(\mvc{x}_T) = \left\| \mvc{p}^L_T - \mvc{p}^L_\text{ref} \right\|^2_{\matSym{P}^L_T} + \left\| \mvc{p}^F_T - \mvc{p}^F_\text{ref} \right\|^2_{\matSym{P}^F_T}
,\end{equation} 
where $\|\mvc{x}\|_{\matSym A} \triangleq \sqrt{\mvc{x}\T \matSym{A} \mvc{x}}$. 
Reference positions $\mvc{p}^i_\textrm{ref}$ are derived from the waypoints $\mvc{q}^d_n = \begin{bmatrix}
    x^d_n & y^d_n & \theta^d_n
\end{bmatrix}\T$ for $n=0,1,\ldots$, produced by the global path planner via
\begin{equation}
    \begin{cases}
    \begin{aligned}
        \mvc{p}^L_\textrm{ref} &= \begin{bmatrix}
            x^d + \frac{l}{2} \cos{\theta^d} & y^d + \frac{l}{2} \sin{\theta^d}
        \end{bmatrix}\T\\
        \mvc{p}^F_\textrm{ref} &= \begin{bmatrix}
            x^d - \frac{l}{2} \cos{\theta^d} & y^d - \frac{l}{2} \sin{\theta^d}
        \end{bmatrix}\T
    .\end{aligned}
    \end{cases}
\end{equation}
For stabilizing the formation of the robot team in order to keep the trolley stack's integrity, we design the stage cost as
\begin{equation}
    J_\text{S}(\mathbf{x}_k) = 
        \lambda^r_k \left|r^2_k - l^2 \right|^2 + 
        \lambda^\phi_k \left( \phi^F_k \right)^2 + 
        \|\mvc{u}_k\|^2_{\matSym{R}_k} + 
        w_k \varepsilon^2_k . 
\end{equation}
The first term intends to stabilize the distance to the desired value and contain the follower robot's orientation to line up with the trolley array. 
The second term penalizes control efforts and the last term punishes the slack on the distance error. 
All coefficients $\matSym{P}^i_T,\, \matSym{R}_k,\, \lambda^r_k,\, \lambda^\phi_k,\,$, and $w_k$ are positive(-definite).

\subsection{Global Path Planning}%
\label{sub:path_planning} 
It is necessary to equip our system with a global path planner because the system is supposed to operate at potentially large-scale
venues like international airports. 
To guarantee kinematic feasibility, we first model the robot-trolley assembly as a car-like virtual vehicle and then find an obstacle-free path given a 2D occupancy map of the operating environment. 
In this work, we define the reference path $\mathcal{D} \subset \mathbb{R}^3$ as a series of waypoints $\mvc{q}^d_n \triangleq \begin{bmatrix}
    x^d_n & y^d_n & \theta^d_n
\end{bmatrix}\T$ for $n=0,1,\ldots$, representing desired planar positions and global orientations of the robot-trolley assembly. 

The choice of the path planner is flexible, and we employ the Hybrid A$^\ast$ planner \cite{dolgov2010path} for simplicity. 
The adopted kinematics adapted from a bicycle model is as follows 
\begin{equation}
\label{eq:ext_bicycle}
    \begin{cases}
        \begin{aligned}
            x_{k+1} &= x_k + v \dtSample \cos(\theta_k +\beta_k) \\
            y_{k+1} &= y_k + v \dtSample \sin(\theta_k +\beta_k) \\
            \theta_{k+1} &= \theta_k + \dfrac{v\cos \beta_k}{l} \dtSample \left( \tan \phi^L_k - \tan \phi^F_k  \right) \\
            \beta_k &= \arctan \left( \frac{1}{2} (\tan \phi^L_k + \tan \phi^F_k  ) \right) ,  
        \end{aligned} 
    \end{cases} 
\end{equation} 
where $\mvc q = \begin{bmatrix}
    x & y & \theta 
\end{bmatrix}\T$ is the 2D pose of a chosen reference point attached on the robot-trolley assembly, 
$v$ denotes the tangent velocity, 
$\phi^L$ and $\phi^F$ are the front and rear ``steers'' of the assembly, respectively, 
and $l$ is the desired distance between the two robots. 
To increase searching speed, we hold the tangent velocity $v$ as a reasonable constant, and let the planner uniformly sample feasible steer control $[\phi^L, \phi^F]$ to extend the virtual vehicle's poses between searching grids. 
The resulting path will serve as a sequence of waypoints and be sent to the motion planning module.

    \subsection{Behavior Selector}%
    \label{sub:fsa} 
    
    \begin{figure}[!tb]
        \centering
        \includegraphics[width=.8\linewidth]{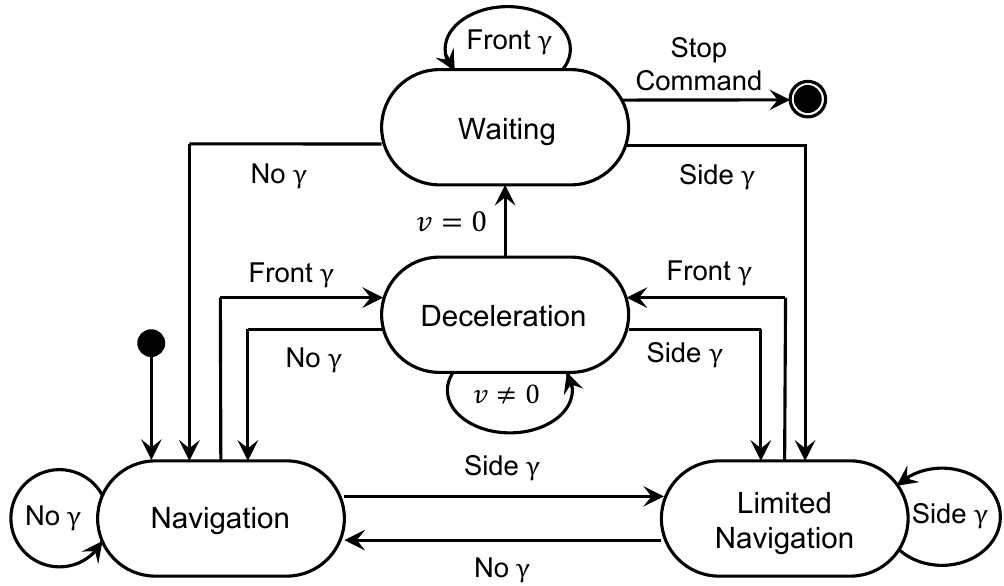}
        \caption{
            Finite-state automaton diagram of the behavior selector module. $\gamma$ represents the detected obstacle which is defined as 3 types by its position and determines the transitions between the 4 modes. The FSA starts in the \emph{Navigation} mode and ends in the \emph{Waiting} mode.
        }
        \label{fig:fsm}
        \vspace{-0.6cm}
    \end{figure}
    
    Operating in densely populated environments like international airports, the robotic system needs to ensure safety and must not bump into people. 
    We identify that it is clumsy, unsafe, and even intimidating for the robots carrying a long array of trolleys to actively perform aggressive maneuvers for obstacle avoidance in such environments. 
    In light of this finding, we propose a \gls{fsa} to manage the robots' behaviors. 
    
    Concretely, we categorize the robots' behaviors in four modes--- \emph{Navigation}, \emph{Deceleration}, \emph{Waiting}, and \emph{Limited Navigation}--- and each mode consists of a state in the proposed \gls{fsa} as shown in \figref{fig:fsm}. 
    The transition between states depends on the relative position between the leader robot and detected dynamic obstacles in its vicinity. 
    The \gls{roi} for dynamic obstacles detection covers an area $\mathcal{O} = \left\{ \gamma = (\rho, \delta) \,\left|\, \rho \in \left[\underline{\rho}, \overline{\rho}\right],\, \delta \in \left[\frac{\pi}{3}, \frac{5\pi}{3}\right] \right. \right\}$, where $\gamma = (\rho, \delta)$ is the polar coordinate of a detected dynamic obstacle with the origin on the leader. 
    When there is no dynamic obstacles in the \gls{roi}, \ie, $\forall \gamma \notin \mathcal{O}$, the robots are in the \emph{Navigation} mode and the motion planner works as default. 
    If obstacles are in the front blocking the robots' way, that is, $\exists \gamma \in \mathcal{O}_1 = \{\gamma \,|\,  \delta \in \left[ \frac{2\pi}{3}, \frac{4\pi}{3} \right] \}$, the robots transit to the \emph{Deceleration} mode and smoothly slow down for a stop. 
    As the robots' velocities gradually decrease to a neighborhood of $0$, the robots are in the \emph{Waiting} state, making way for surrounding dynamic obstacles like humans. 
    The \emph{Limited Navigation} mode is activated for smooth braking and starting when the robots perceive humans on either side of the robot-trolley assembly ($\exists \gamma \in \mathcal{O}_2 = \mathcal{O}\backslash \mathcal{O}_1 \land \forall \gamma \notin \mathcal O_1$). 
    In this mode, the robots move with a lower speed limit. 
    The robots will switch back to \emph{Navigation} if the detected humans move away, or enter \emph{Deceleration} if any person moves from the side to the front.

    \subsection{State Estimation}
    \label{sub:state_estimation}
    
    It is of paramount importance to have reliable information on the robots' states due to the demand for high motion precision for formation maintenance in this transportation task.
    As such, aside from using LiDARs, IMUs, and odometry for the localization of each individual robot, we have added markers on the leader robot so that we can measure the relative pose between the two robots from the follower with its LiDAR readings. 
    
    We adopt an \gls{ekf} to fuse partial information coming from different sources: the localization modules on two robots and the relative pose measurement. 
    The system's discrete state-space representation is: 
    \begin{equation}
    \label{eq:concatenated_dyn}
        \begin{cases}
        \begin{aligned}
            \mathbf{x}_{k+1} &= f(\mathbf{x}_k, \mathbf{u}_k) = \mathbf{x}_k + \matSym{G}(\mathbf{x}_k) \mathbf{u}_k + \mathbf{v}_k\\
            \mathbf y^L_{k+1} &= \matSym{H}_L \mathbf x_{k+1} + \mathbf{w}^L_{k+1}\\
            \mathbf y^F_{k+1} &= \matSym{H}_F \mathbf x_{k+1} + \mathbf{w}^F_{k+1}\\
            \mathbf y^{\Delta}_{k+1} &= \matSym{H}_\Delta \mathbf x_{k+1} + \mathbf{w}^\Delta_{k+1},
        \end{aligned}
        \end{cases}
    \end{equation}
    where 
    \begin{align*}
        \matSym{H}_L \triangleq 
            \begin{bmatrix}
                \matSym{I}_3 & \matSym{0}
            \end{bmatrix},\; 
        \matSym{H}_F \triangleq 
            \begin{bmatrix}
                \matSym{0} & \matSym{I}_3
            \end{bmatrix},\; 
        \matSym{H}_\Delta \triangleq 
            \begin{bmatrix}
                \matSym{I}_3 & -\matSym{I}_3
            \end{bmatrix}.
    \end{align*}
    $\mvc{y}^i_k \in \mathbb R^3$ for $i\in\{L,F\}$ are pose measurements obtained from the robots' localization modules; $\mvc y^\Delta_k \in \mathbb R^3$ is a relative pose measurement. 
    We assume that all noises follow zero-mean Gaussian distribution and are independent of one another, \ie, $\mathbf{v}_k \sim \mathcal N(\mvc{0}, \matSym{V}_k)$, $\mathbf{w}^i_k \sim \mathcal N(\mvc{0}, \matSym{W}^i_k)$ for $i\in \{L, F, \Delta\}$. 
    Since we perform local motion planning in a centralized fashion, the prediction step is standard as that of any \gls{ekf}.  
    Measurements update at different frequencies, so we perform update steps asynchronously whenever an observation comes in. 
    In practice, we find this fusion of multi-source information critical to the smooth functioning of the motion planner and facilitative of highly cooperative operation of the robots. 
    
    

%% file: sections/5_exp.tex
\section{Implementation and Experiments}
\label{sec:experiments}
We present experimental results to validate the effectiveness of the proposed collaborative motion planning module and demonstrate our trolley transportation system in complex and dynamic environments.

    \subsection{Implementation Details}
        \subsubsection{Robot Setup} 
        We use two autonomous robots to perform collaborative trolley transportation as mentioned in \secref{sub:design}. 
        Our framework is implemented in Python and integrated through Robot Operating System (ROS Melodic). 
        Each robot is equipped with an Intel NUC (specs: Core i7-1165G7 CPU@4.70GHz, 32GB RAM) to perform all computations online. 
        Onboard sensors include an Ouster OS1 128-channel LiDAR and a 9-axis IMU.
        Both robots communicate through a wireless ad-hoc network with their onboard WiFi routers. 
        
        \subsubsection{Localization}
        Each robot runs its own global localization via AMCL \cite{fox2001kld} with a 2D occupancy map built \apriori using GMapping \cite{gmapping}. 
        For high-fidelity measurements of the robots' relative position, we install a cylinder marker with unique reflectivity at the leader's rotation center. 
        The follower then extracts the marker's position from the LiDAR scan points and takes it as the leader's position.

        \subsubsection{Planning}
        The global path planner is set up with $0.25\si{m} \times 0.25\si{m} \times 15^\circ$ resolution, and the virtual vehicle's size is slightly expanded for safety. 
        The motion planner runs at approximately $30\si{Hz}$ with a horizon of $T=20$ steps, each discretized by $\dtSample=0.1\si{s}$. 
        The MPC problem is formulated by CasADi \cite{andersson2019casadi} and solved with IPOPT \cite{biegler2009large} on one robot's onboard computer in real-time.

    \begin{figure}[tb]
        \centering
        \includegraphics[width=0.98\linewidth]{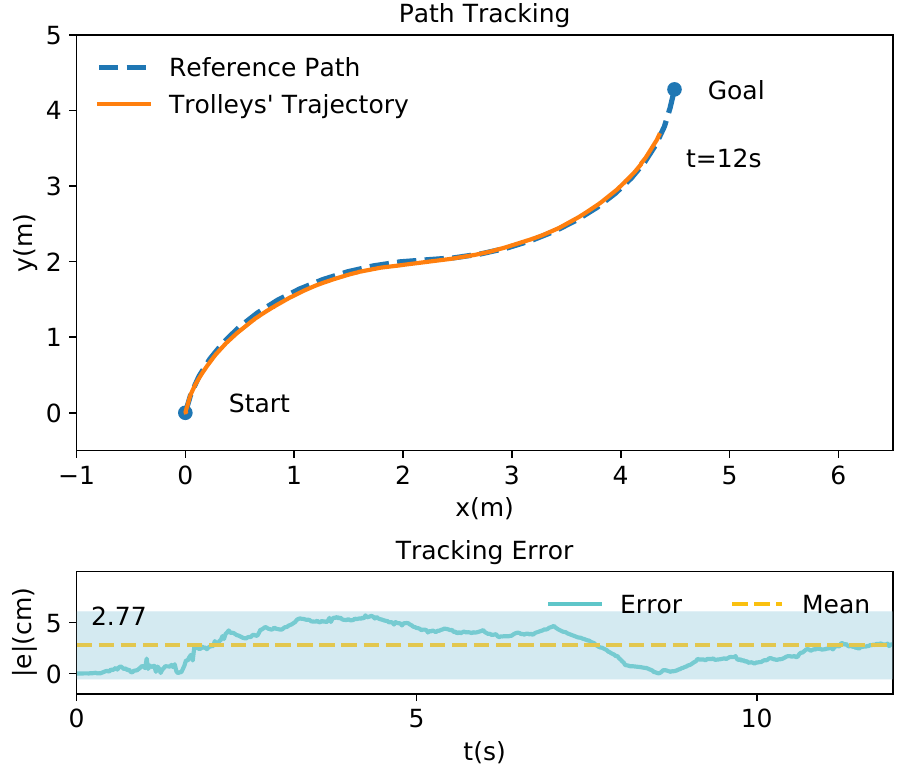}
        \caption{
        Results of the path tracking performance evaluation. 
        The upper figure delineates the top view of the reference path and the trajectory of the reference point on the robot-trolley assembly till $t=12\,\si{s}$. 
        The lower figure plots the tracking error and related statistics over time. 
        The shaded area represents the error's $2\sigma$-margin.}
        \label{fig:exp_in}
        \vspace{-5mm}
    \end{figure}

    \subsection{Collaborative Motion Planning Evaluation}
    To test the tracking performance of our collaborative motion planning method, we conduct a real-world experiment transporting three trolleys to track a given reference path in a motion capture area without obstacles. 
    An OptiTrack motion capture system is used to provide precise localization feedback and validation. 
    The reference path consists of 30 waypoints sampled from a concatenation of two arcs both with a $0.44\,\si{m}^{-1}$ curvature.

    The tracking results are shown in \figref{fig:exp_in}. 
    In this experiment, we focus on the $X$-$Y$ position of a reference point attached on the robot-trolley assembly (in orange). 
    The experiment starts at $(0,0)$ and the robots transport trolleys along the predefined reference path to the goal region with a center at $(4.5, 4.2)$. 
    The two robots accomplish the collaborative transportation task in $12.2\si{s}$ with an average velocity of $0.491\si{m/s}$, and stop when the distance to the goal is within $0.3\si{m}$. 
    Along the entire way, the tracking error is $2.77\pm 1.68\si{cm}$ with a maximal error of $5.66\si{cm}$.
    \footnote{The tracking error is defined as the Euclidean distance between the two curves. For tracking error calculation only, the reference path which contains 30 waypoints is densified via cubic interpolation.}

\begin{figure}[tb]
    \centering
    \includegraphics[width=0.95\linewidth]{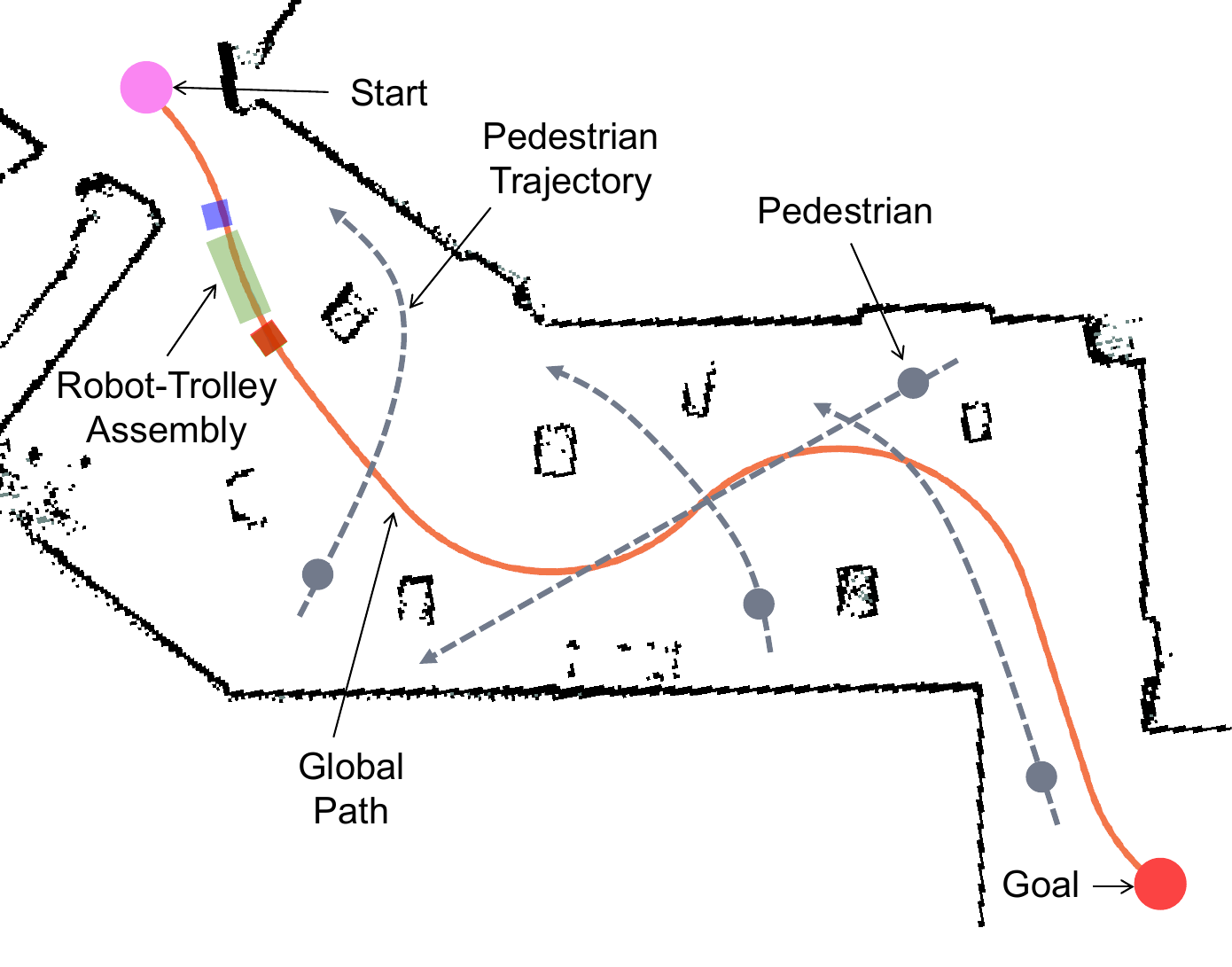}
    \caption{
    A top-view visualization illustrating our experiment setup. 
    The global path is painted orange. 
    Static obstacles in the occupancy map are represented by black points. 
    The gray cylinders (circles in this figure) represent detected pedestrians and grey dotted lines denote their trajectories. 
    The robots autonomously find the global path (orange curve) of the whole assembly and plan their own motions. 
    }
    \label{fig:setup}
    \vspace{-5mm}
\end{figure}

\begin{figure*}[tb!]
    \centering
    \subfigure[Experiment of collaborative transportation of five trolleys in a cluttered narrow space.]
    {\includegraphics[width=0.97\linewidth]{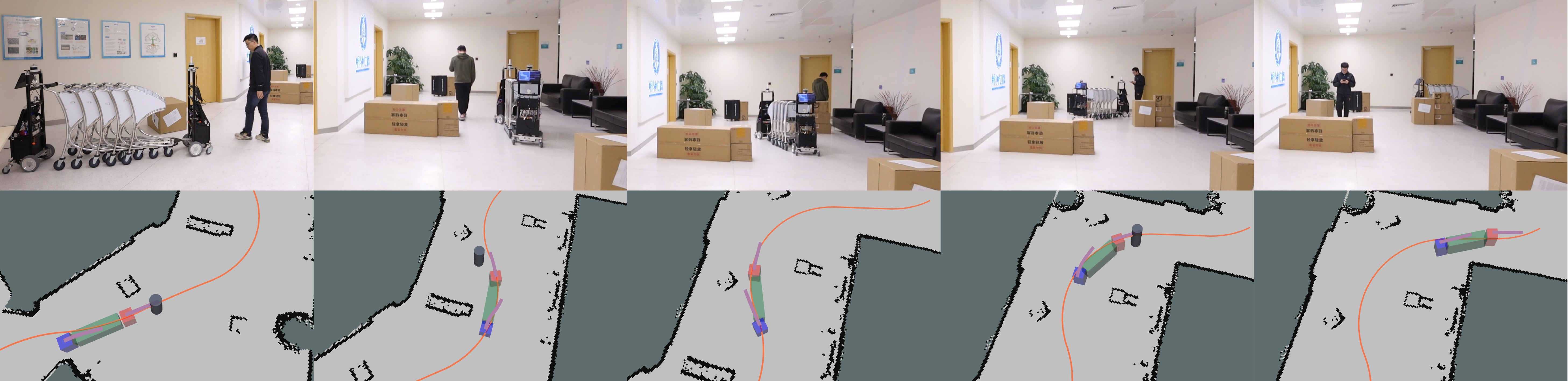}\label{fig:exp1}}\\
    \vspace{-2mm}
    \subfigure[Robots' velocities, bearing angles, and the relative distance error in the cluttered narrow space.]{\includegraphics[width=0.97\linewidth]{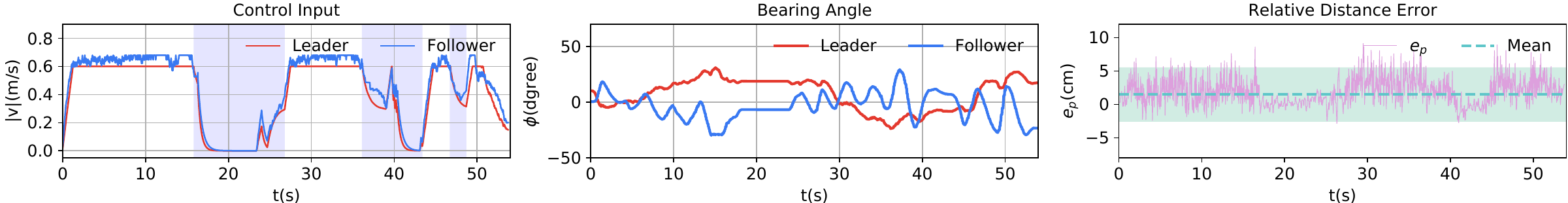}\label{fig:exp2}}\\
    \vspace{-2mm}
    \subfigure[Experiment of collaborative transportation of eight trolleys in presence of more pedestrians around.]{\includegraphics[width=0.97\linewidth]{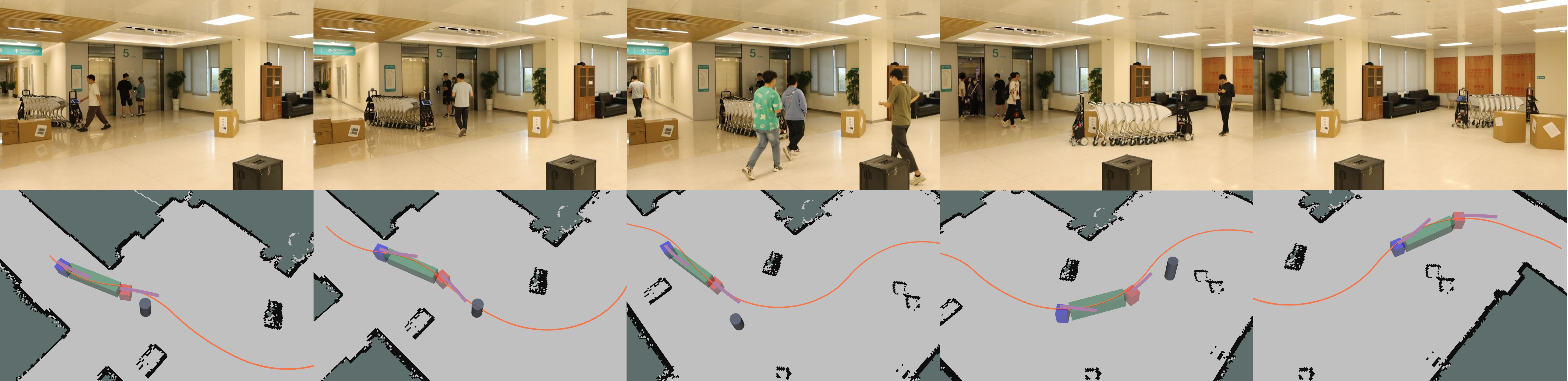}\label{fig:exp3}}\\
    \vspace{-2mm}
    \subfigure[Robots' velocities, bearing angles, and the relative distance error in the populated space.]{\includegraphics[width=0.97\linewidth]{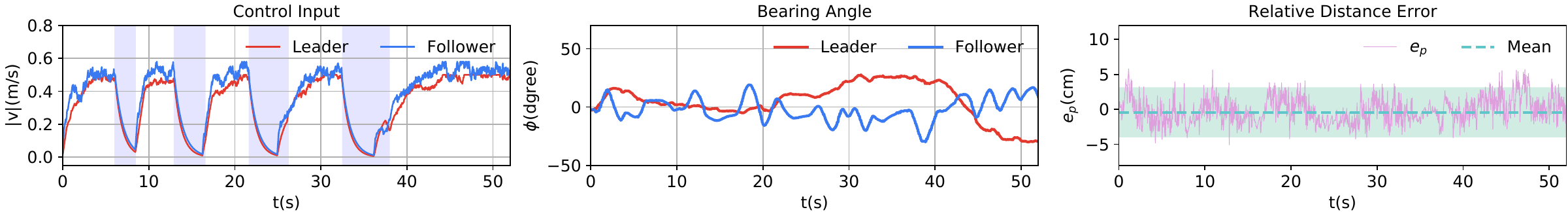}\label{fig:exp4}}
    \vspace{-2mm}
    \caption{
    Snapshots and visualization of the proposed system working in two typical scenarios: narrow spaces and crowded spaces. 
    In both scenarios, the visualization shows the robot-trolley assembly (green, blue, and red cubes), the global path (orange), the local motion plans of each robot (light purple), and the nearest detected pedestrian (gray cylinder). 
    Three metrics of interest were also recorded: robots' linear velocities, bearing angles, and the relative distance error. 
    The light blue shades denote times when the behavior selector takes effect to yield for humans; light green shaded areas indicate the distance errors' $2\sigma$-margins. 
    The proposed framework clearly empowered our system to safely and smoothly perform collaborative trolley transportation tasks. 
    }
    \label{fig:exp}
    \vspace{-5mm}
\end{figure*}

    \subsection{Trolley Transportation Demonstration}    
    To demonstrate the practical applicability of our proposed system for trolley transportation in real-world environments such as airports or supermarkets, we conduct experiments in complex and dynamic scenarios where the robots need to transport multiple trolleys while navigating through cluttered static obstacles and pedestrians. 
    \figref{fig:setup} illustrates the graphic representation of one experiment's setup. 
    As a result, the snapshots, the relevant data, and the visualization of the experiments are shown in \figref{fig:exp}.

        \subsubsection{Narrow Space}
        In this scenario, the robots are collaboratively transporting 5 trolleys across a narrow space stuffed with 6 static obstacles and 3 pedestrians. 
        The trolley stack is $1.98\,\si{m}$ long and weighs around $100\,\si{kg}$. 
        It is not easy even for an adult to manually handle a queue of more than five trolleys in such a complex environment. 

        As expected, the path planner successfully identifies a smooth global path, which is denoted as the orange curve in the visualization in \figref{fig:exp1}. 
        Although the obstacles result in a serpentine path with several turns almost to the robot-trolley assembly's kinematic limit, the robots operate around their full speeds $v^F_{\text{max}} = 0.7\,\si{m/s}$, $v^L_{\text{max}} = 0.6\,\si{m/s}$ when the path is clear of dynamic obstacles. 
        As pedestrians get in the way, the proposed behavior selector kicks in so the robots smoothly stop, firmly hold the trolleys for safety, and resume moving when the path gets clear, just as is shown in the shaded areas of the velocity plot in \figref{fig:exp2}, starting at time $t \in \{15.8\si{s}, 36\si{s}, 46\si{s}\}$, respectively. 
        
        Another critical requirement for successful transportation is to maintain the trolley array's integrity, which is reflected by the relative pose of the two robots. 
        Since the robots' manipulators are not rigidly fixed on the trolleys, large variations in the relative distance or orientation difference between robots will cause detachment or structural damage. 
        Deriving from the relative pose measured by the proposed LiDAR-marker detection method, \figref{fig:exp2} shows the error to the desired distance $e_p = 1.49\pm 2.03 \,\si{cm}$ and the difference between the orientation of the trolley stack and those of both robots are well kept within the range of $\pm 45^\circ$. 

        \subsubsection{Populated Space}
        This scenario involves two robots transporting 8 trolleys with a total length of $2.56\,\si{m}$ in a broad hall filled with more moving humans. 
        The velocity limits are set as $v^F_{\text{max}} = 0.58\,\si{m/s}$, $v^L_{\text{max}} = 0.5\,\si{m/s}$, and the acceleration is smaller. 
        As presented in \figref{fig:exp3}, the robot-trolley assembly successfully avoids all obstacles, makes way for several humans, tracks the reference path, and reaches the goal position. 
        The fluctuations in the follower's velocity depicted in \figref{fig:exp4} indicate the adjustment to keep the trolleys stuck tight. 
        The relative distance error $e_p$ to the desired distance is $0.255 \pm 2.36\,\si{cm}$. 
        \footnote{The range accuracy of the Ouster OS1 LiDARs on the robots is $2.5\,\si{cm}$.}

        These two experiments manifest the effectiveness of the proposed multi-robot system and demonstrate that our autonomy framework can provide highly cooperative motions despite a variety of underlying complex constraints for safe and smooth operation in collaborative trolley transportation.

%% file: sections/6_conclusion.tex
\section{Conclusions and Future Work}
\label{sec:conclusion}
This paper presents a hierarchical navigation framework and integrates it into an autonomous multi-robot system designed for collaborative trolley transportation. 
By dividing the various inherent constraints and tackling them at different decision-making levels, our framework is robust and fast enough to provide real-time solutions for the challenging collaborative trolley transportation task. 
We demonstrate the proposed system to achieve cooperative trolley transportation in different scenarios. 
Experimental results have exhibited the efficacy and reliability of the proposed method and hardware system in cluttered and populated dynamic environments. 

In the future, we will proceed with socially-aware motion planning and more large-scale outdoor realistic experiments. 
Moreover, we shall actively explore decentralized solutions of our framework to improve robustness and scalability.